\documentclass[runningheads,a4paper]{llncs}

\usepackage{amssymb}
\usepackage{amsmath}
\usepackage{graphicx}
\usepackage[table,pdftex]{xcolor}
\usepackage{xspace}
\usepackage[colorlinks=true,citecolor=blue,urlcolor=black]{hyperref}
\usepackage{lineno}
\usepackage{subcaption} 
\usepackage[colorinlistoftodos]{todonotes}

\usepackage{booktabs}
\usepackage{array}
\usepackage[flushleft]{threeparttable}

\begin{document}

\mainmatter  
\title{Ensembles of Multiple Models and Architectures for Robust Brain Tumour Segmentation}
\titlerunning{EMMA for Robust Brain Tumour Segmentation}

\author{K. Kamnitsas, W. Bai$^*$, E. Ferrante$^*$, S. McDonagh$^*$, M. Sinclair$^*$\\
N. Pawlowski, M. Rajchl, M. Lee, B. Kainz, D. Rueckert, B. Glocker}
\institute{Biomedical Image Analysis Group, Imperial College London, UK\\
$^*$ Equal contribution, in alphabetical order}
\authorrunning{Kamnitsas, Bai, Ferrante, McDonagh, Sinclair, et al.}
\maketitle



\begin{abstract}

Deep learning approaches such as convolutional neural nets have consistently outperformed previous methods on challenging tasks such as dense, semantic segmentation. However, the various proposed networks perform differently, with behaviour largely influenced by architectural choices and training settings. This paper explores Ensembles of Multiple Models and Architectures (EMMA) for robust performance through aggregation of predictions from a wide range of methods. The approach reduces the influence of the meta-parameters of individual models and the risk of overfitting the configuration to a particular database. EMMA can be seen as an unbiased, generic deep learning model which is shown to yield excellent performance, winning the first position in the BRATS 2017 competition among 50+ participating teams.

\end{abstract}


\section{Introduction}

Brain tumours are among the most fatal types of cancer \cite{deangelis2001brain}. Out of tumours that originally develop in the brain, gliomas are the most frequent \cite{bauer2013survey}. They arise from glioma cells and, depending on their aggressiveness, they are broadly categorized into high and low grade gliomas \cite{louis20162016}. High grade gliomas (HGG) develop rapidly and aggressively, forming abnormal vessels and often a necrotic core, accompanied by surrounding oedema and swelling \cite{bauer2013survey}. They are malignant, with high mortality and average survival rate of less than two years even after treatment \cite{louis20162016}. Low grade gliomas (LGG) can be benign or malignant, grow slower, but they may recur and evolve to HGG, thus their treatment is warranted. For treatment, patients undergo radiotherapy, chemotherapy and surgery \cite{deangelis2001brain}.

Firstly for diagnosis and monitoring the tumour's progression, then for treatment planning and afterwards for assessing the effect of treatment, various neuro-imaging protocols are employed. Magnetic resonance imaging (MRI) is widely used in both clinical routine and research studies. It facilitates tumour analysis by allowing estimation of extent, location and investigation of its subcomponents \cite{bauer2013survey}. This however requires accurate delineation of the tumour, which proves challenging due to its complex structure and appearance, the 3D nature of the MR images and the multiple MR sequences that need to be consulted in parallel for informed judgement. These factors make manual delineation time-consuming and subject to inter- and intra-rater variability \cite{menze2015multimodal}.

Automatic segmentation systems aim at providing an objective and scalable solution. Representative early works are the atlas-based outlier detection method \cite{prastawa2004brain} and the joint segmentation-registration framework, often guided by a tumour growth model \cite{gooya2011joint,parisot2012joint,bakas2015glistrboost}. The past few years saw rapid developments of machine learning methods, with Random Forests being among the most successful \cite{zikic2012decision,le2016lifted}. More recently, convolutional neural networks (CNN) have gained popularity by exhibiting very promising results for segmentation of brain tumours \cite{urban2014CnnBrats,pereira2016brain,kamnitsas2017efficient}.

\begin{figure}[t] 
\centering
\begin{subfigure}[b]{0.225\textwidth}
	\centering
	\includegraphics[clip=true, trim=0pt 0pt 0pt 0pt, width=1.\textwidth]{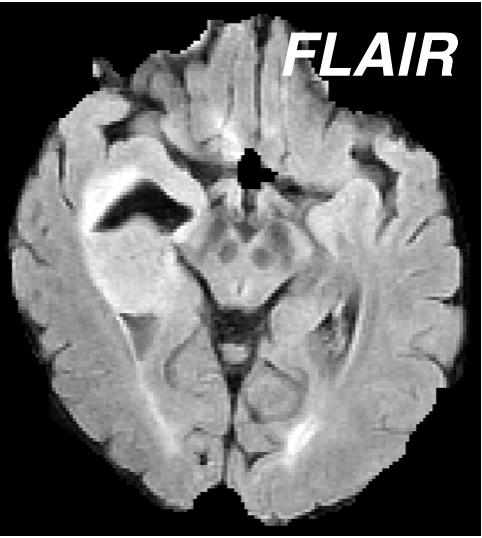}
	\label{fig:flair}
\end{subfigure}
\begin{subfigure}[b]{0.225\textwidth}
	\centering
	\includegraphics[clip=true, trim=0pt 0pt 0pt 0pt, width=1.\textwidth]{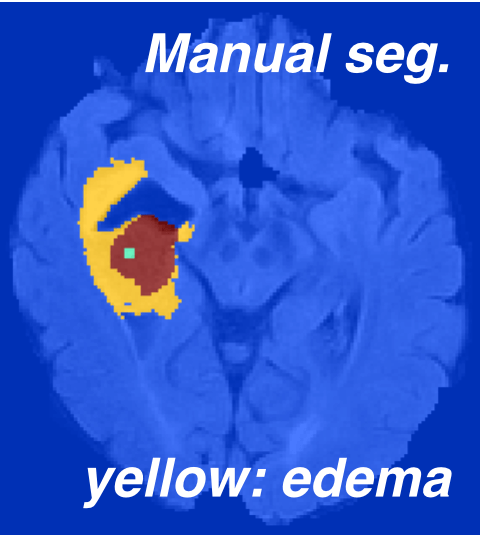}
	\label{fig:gt}
\end{subfigure}
\begin{subfigure}[b]{0.225\textwidth}
	\centering
	\includegraphics[clip=true, trim=0pt 0pt 0pt 0pt, width=1.\textwidth]{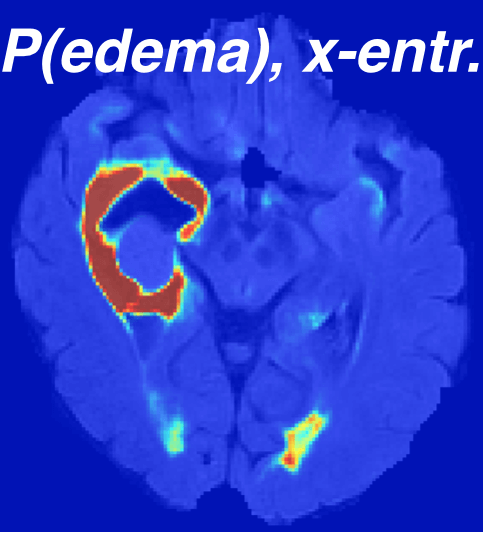}
	\label{fig:xEntr}
\end{subfigure}
\begin{subfigure}[b]{0.225\textwidth}
	\centering
	\includegraphics[clip=true, trim=0pt 0pt 0pt 0pt, width=1.0\textwidth]{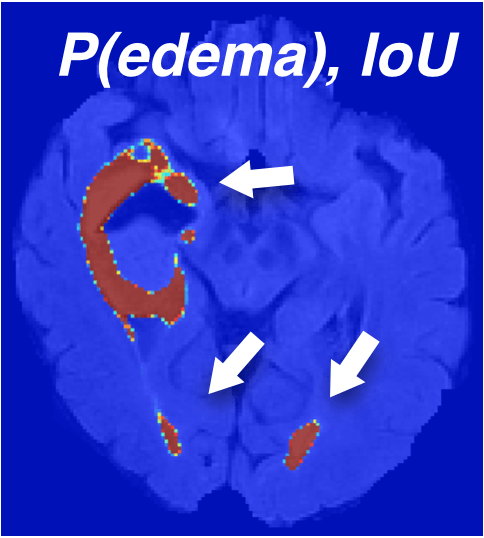}
	\label{fig:iou}
\end{subfigure}
\vspace{-5mm}
\caption{Left to right: FLAIR; manual annotation of a BRATS'17 subject, where yellow depicts oedema surrounding tumour core; confidence of a CNN predicting oedema, trained with cross-entropy or IoU loss. Although overall performance is similar, training with IoU (or Dice, not shown) loss alters the CNN's behaviour, which tends to output only highly confident predictions, even when false.}
\label{fig:xEntrVsIou}
\end{figure}

A variety of CNN architectures have been proposed, each presenting different strengths and weaknesses. Additionally, networks have a vast number of meta parameters. The multiple configuration choices for a system influence not only performance but also its behaviour (Fig.~\ref{fig:xEntrVsIou}). For instance, different models may perform better with different types of pre-processing. Consequently, when investigating their behaviour on a given task, findings can be biased. Finally, a configuration highly optimized on a given database may be an over-fit, and not generalise to other data or tasks.

In this work we push towards constructing a more \textit{reliable} and \textit{objective} deep learning model. We bring together a variety of CNN architectures, configured and trained in diverse ways in order to introduce high variance between them. By combining them, we construct an \textit{Ensemble of Multiple Models and Architectures} (EMMA), with the aim of \textit{averaging away} the variance and with it model- and configuration-specific behaviours. Our approach leads to: (1) a system robust to unpredictable failures of independent components, (2) enables objective analysis with a generic deep learning model of unbiased behaviour, (3) introduces the new perspective of \textit{ensembling for objectiveness}. This is in contrast to common ensembles, where a single model is trained with small variations such as initial seeds, which renders the ensemble biased by the main architectural choices. As a first milestone in this endeavour, we evaluated EMMA in the Brain Tumour Segmentation (BRATS) challenge 2017. Our method won the first position in the final testing stage among 50+ competing teams. This indicates the reliability of the approach and paves the way for its use in further analysis.


\section{Background: Model bias, variance and ensembling}
\label{sec:related_work}

Feedforward neural networks have been shown capable of approximating any function \cite{hornik1989multilayer}. They are thus models with zero bias, possible of no systematic error. However they are not a panacea. If left unregularized they can overfit noise in the training data, which leads to mistakes when they are called to generalise. Coupled with the stochasticity of the optimization process and the multiple local minima, this leads to unpredictable inconsistent errors between different instances. This constitutes models with high variance. Regularization reduces the variance but increases the bias, as expressed in the bias/variance dilemma \cite{geman2008neural}. Regularization can be explicit, such as weight decay that prevents networks from learning rare noisy patterns, or implicit, such as the local connectivity of CNN kernels, which however does not allow the model to learn patterns larger than the its receptive field. Architectural and configuration choices thus introduce bias, altering the behaviour of a network.

One route to address the bias/variance dilemma is ensembling. By combining multiple models, ensembling seeks to create a higher performing model with low variance. The most popular combination rule is averaging, which is not sensitive to inconsistent errors of the singletons \cite{kittler1998combining}. Commonly, instances of a network trained with different initial weights or from multiple final local minima are ensembled, with the majority correcting irregular errors. Intuitively, only inconsistent errors can be averaged out. Lack of consistent failures can be interpreted as statistical independency. Thus methods for de-correlating the instances have been developed. The most popular is \textit{bagging} \cite{breiman1996bagging}, commonly used for random forests. It uses bootstrap sampling to learn less correlated instances from different subsets of the data.

The above works often discuss ensembling as a means of increasing performance. \cite{sharkey1997combining} approached high variance from the scope of \textit{unreliability}. They discussed ensembling as a type of N-version programming, which advocates reliability through redundancy. When producing N-versions of a program, versions may fail independently but through majority voting they behave as a reliable system. They formalize intuitive requirements for reliability: a) the target function to be covered by the ensemble and b) the majority to be correct. This in turn advocates diversity, independence and overall quality of the components.

Biomedical applications are reliability-critical and high variance would deter the use of neural networks. For this reason we set off to investigate robustness of diverse ensembles. Diverting from the above works, we introduce another perspective of ensembling: creating an objective, configuration-invariant model to facilitate objective analysis.


\section{Ensembles of Multiple Models and Architectures}
\label{sec:method}

\begin{figure}[t] 
\centering
\includegraphics[clip=true, trim=0pt 0pt 0pt 0pt, width=1.\textwidth]{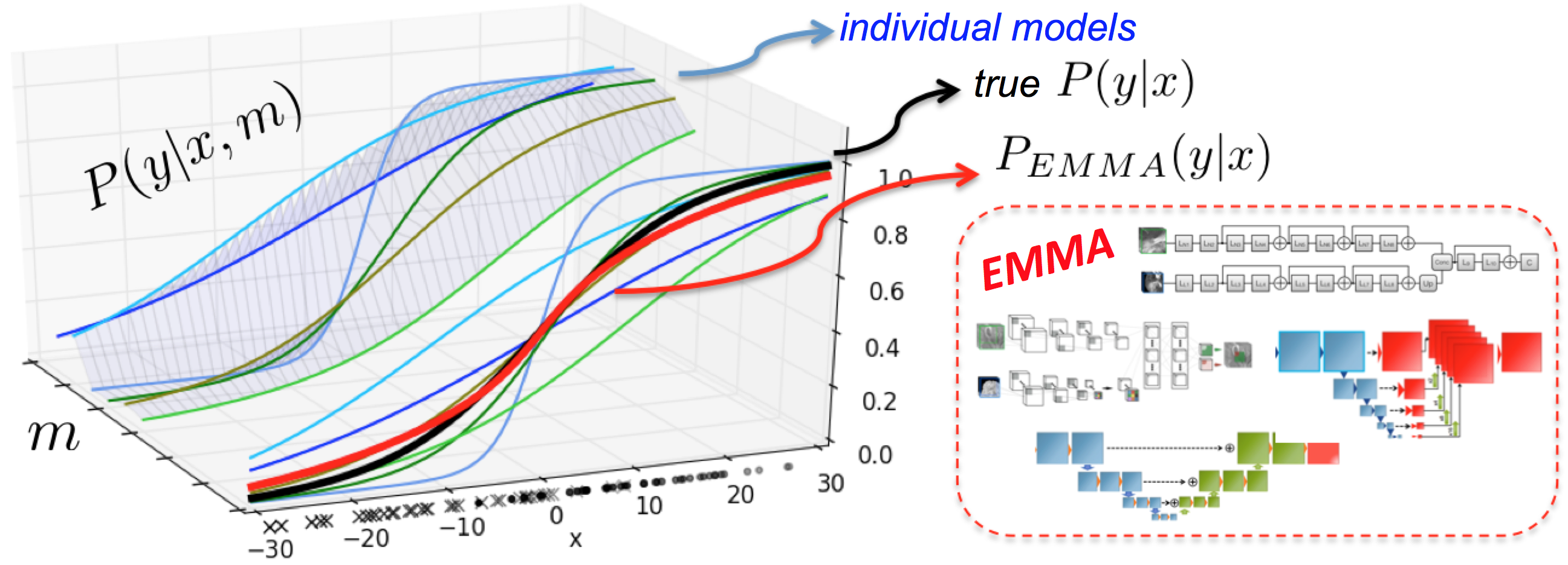}
\caption{Our ensemble of diverse networks, EMMA (red), averages out the bias infused by individual model configurations $m$, to approximate more reliably the true posterior (black), while being robust to suboptimal configurations. Posteriors on the left were obtained from multiple perceptrons, trained to classify clusters centred on 10 and -10 as a toy example, with different losses, regularizations and noise in the training labels. Their ensemble provides reliable estimates.}
\label{fig:mainFig}
\end{figure}

A variety of CNN architectures has shown promising results in recent literature. Regarding the architectures, they commonly differ in depth, number of filters and how they process multi-scale context among others. Such architectural choices bias the model behaviour. For instance, models with large receptive fields may show improved localisation capabilities but can be less sensitive to fine texture than models emphasizing local information. Strategies to handle class imbalance is another performance relevant parameter. Common strategies are training with class-weighted sampling or class-weighted cross entropy. As analysed in \cite{kamnitsas2017efficient}, these methods strongly influence the sensitivity of the model to each class. Furthermore, the choice of the loss function impacts results. For example, we observed that networks trained to optimize Intersection over Union (IoU), Dice or similar losses \cite{nowozin2014optimal} tend to give worse confidence estimations than when trained with cross entropy (Fig.~\ref{fig:xEntrVsIou}). Finally, the setting of hyper-parameters for the optimization can strongly affect performance. It is often observed by practitioners that the choice of the optimizer and its configuration, for instance the learning rate schedule, can make the difference between bad and good segmentation.

The sensitivity to such meta-parameters is a greater problem than merely a time-consuming manual optimization of configurations:
\begin{itemize}
\item A configuration setting optimized on one set of training data may be over-fitting them and not perform well on unseen data or another task. This can be viewed as another source of high model variance (Sec.~\ref{sec:related_work}).
\item By biasing the behaviour of the model, it also biases the findings of any analysis performed with it.
\end{itemize}

We now formalize the problem and our perspective of ensembling as a solution as follows. Given training data $X$ with labels $Y$, we need to learn the generating process $P(y|x)$. This is commonly approximated by a model $P(y|x;\theta_m,m)$, which has trainable parameters $\theta_m$ that are learnt via an optimization process that minimizes:
\begin{equation}
\label{eq:optimization}
\theta_m = \min_{\theta_m} d ( P(Y|X;\theta_m,m), P(Y|X))
\end{equation}
where $d$ is a distance (defined by the type of loss) computed at the points given by the training data, while $m$ represents the choice of the meta-parameters. It is commonly neglected although it conditions (biases) the learnt estimator. To take it into account, we instead define $m$ as a stochastic variable over the space of meta-parameter configurations, with a corresponding prior $P(m)$. In order to learn a model of $P(y|x)$ unbiased by $m$, we marginalize out its effect:
\begin{equation}
\label{eq:emmaFormula}
\begin{split}
P(y|x) = \sum_{m} P( y, m | x ) &= \sum_{m} P(y | x, m ) P(m) \\
&\approx \sum_{\forall{m}\in{E}} P(y|x;\theta_m,m) \frac{1}{|E|} = P_{EMMA}(y|x)
\end{split}
\end{equation}

Here $E$ is the set of models within the ensemble. The prior $P(m)$ is considered uniform over a subspace of $m$ that is covered by the models in $E$ and zero elsewhere. Note we have arrived at the standard ensembling with averaging, by considering that each individual model $P(y|x;\theta_m,m)$ approximates a conditional $P(y|x, m)$ on $m$, and the true posterior is approximated by the ensemble which marginalizes away effects of $m$. Note that the case of a single model configured by $m$ can be derived from the above, by setting a dirac prior $P(m)=\delta(m)$. Thus the ensemble relaxes a pre-existing neglected strong prior.

The above formulation presents averaging ensembles from a new perspective: The marginalization over a subspace of the joint $P(y | x, m )$ offers generalisation, regularising the (manual) optimization process of $m$ from falling into minima where $P(Y | X, m )$ overfits $P(Y|X)$ on the given training data $(Y,X)$ (Fig.~\ref{fig:mainFig}). Moreover, the process leads to a more objective approximation of $P(y|x)$ where the biasing effect of $m$ has been marginalized out. The exposed limitations agree with the requirements for ensembling mentioned in Sec.~\ref{sec:related_work}: we need to restrict the subspace of $m$ into an area of relatively high quality models and we need to cover it with a relatively small number of models, thus diversity is key.

In the remainder of this section we describe the main properties of the models used to construct the collection $E$ of EMMA, which cover various contemporary architectures, configured and trained under different settings\footnote{Implementation and configuration details considered less important for this work were omitted to avoid cluttering the manuscript.}.

\subsection{DeepMedic}

\begin{figure}[t] 
\centering
\includegraphics[clip=true, trim=0pt 0pt 0pt 0pt, width=1.\textwidth]{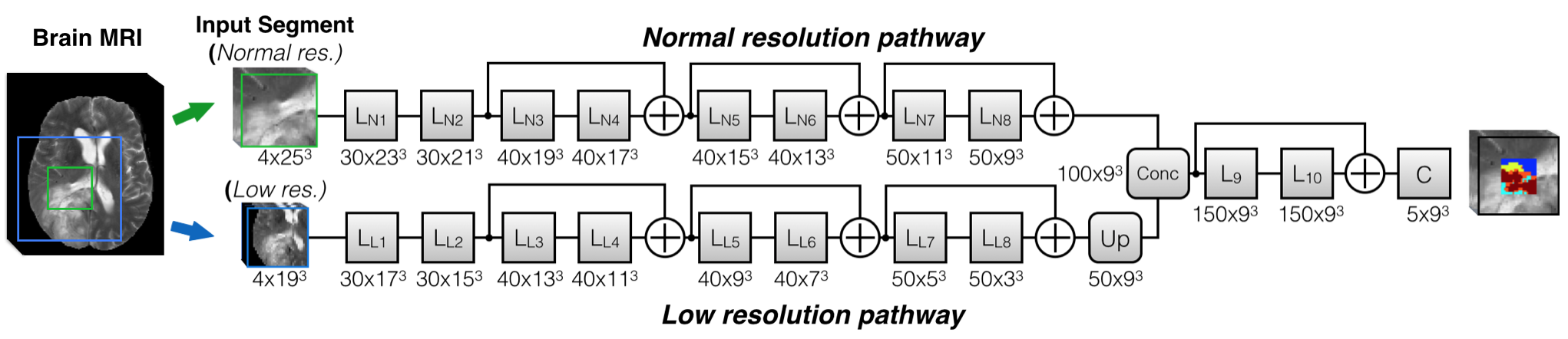}
\caption{We used two DeepMedics \cite{kamnitsas2017efficient} in our experiments. The smaller of the two is depicted, where the number of feature maps and their dimension at every layer are depicted in the format $( Number \times Size)$. The second model used in the ensemble is wider, with double the number of feature maps at every layer. All kernels and feature maps are 3D, even though not depicted for simplicity.}
\label{fig:resDm}
\end{figure}

\noindent \textit{Model description:} The first architecture we employ is DeepMedic, originally presented in \cite{kamnitsas2015Isles,kamnitsas2017efficient}. It is a fully 3D, multi-scale CNN, designed with a focus on efficient processing of 3D images. For this, it employs parallel pathways that take as input down-sampled context, avoiding to convolve large volumes at full resolution to remain computationally cheap. Although originally developed for segmentating brain lesions, it was found promising on diverse tasks, such as segmentation of the placenta \cite{alansary2016fast}, making it a good component for a robust ensemble. We include two deepMedic models in EMMA. The first is the residual version previously employed in BRATS 2016 \cite{kamnitsas2016deepmedic}, depicted in Fig.~\ref{fig:resDm}. The second is a wider variant, with double the number of feature maps at each layer.

\noindent \textit{Training details:} The models are trained by extracting multi-scale image segments with a 50\% probability centred on healthy tissue and 50\% probability on tumour as proposed in \cite{kamnitsas2017efficient}. The wider variant is trained on larger inputs of width 34 and 22 for the two scales respectively. They are trained with cross-entropy loss, with all meta-parameters adopted from the original configuration.

\subsection{FCN}

\begin{figure}[!h] 
\centering
\includegraphics[clip=true, trim=0pt 0pt 0pt 0pt, width=0.9\textwidth]{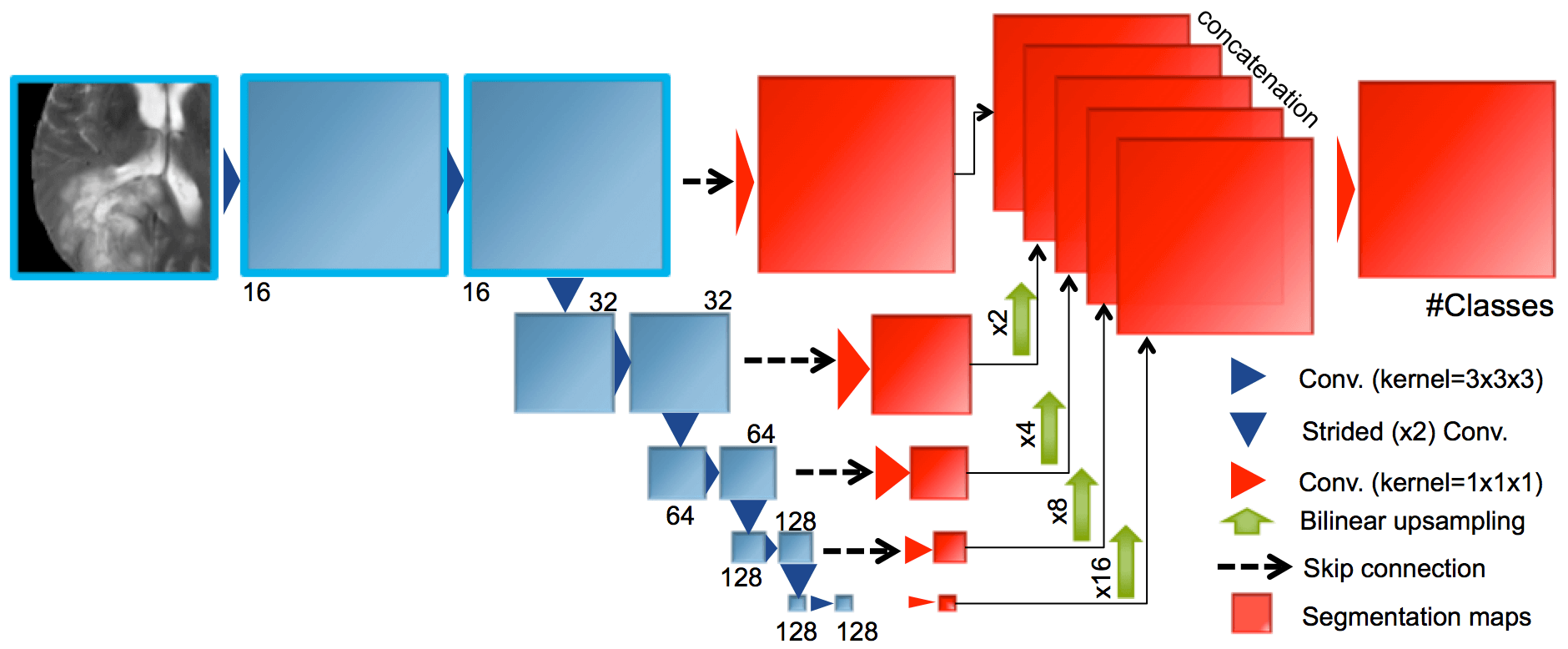}
\caption{Schematic of one of the FCN architecture used in EMMA. Shown are number of feature maps per layer. All kernels and feature maps are 3D, even though not depicted for simplicity.}
\label{fig:fcn}
\end{figure}

\noindent \textit{Model description:} We integrate three 3D FCNs \cite{Long2015} in EMMA. A schematic of the first architecture is depicted in Fig.~\ref{fig:fcn}. The second FCN is constructed larger, replacing each convolutional layer with a residual block with two convolutions. The third is also residual-based, but with one less down-sampling step. All layers use batch normalisation, ReLUs and zero-padding.

\noindent \textit{Training details:} We draw training patches of width 64 for the first and 80 voxels for the residual-based FCNs, with an equal probability from each label. They were trained using Adam. The first was trained to optimize the IoU loss \cite{nowozin2014optimal} while the Dice was used similarly for the other two.

\subsection{U-Net}
\begin{figure}[t] 
\centering
\includegraphics[clip=true, trim=0pt 0pt 0pt 0pt, width=0.9\textwidth]{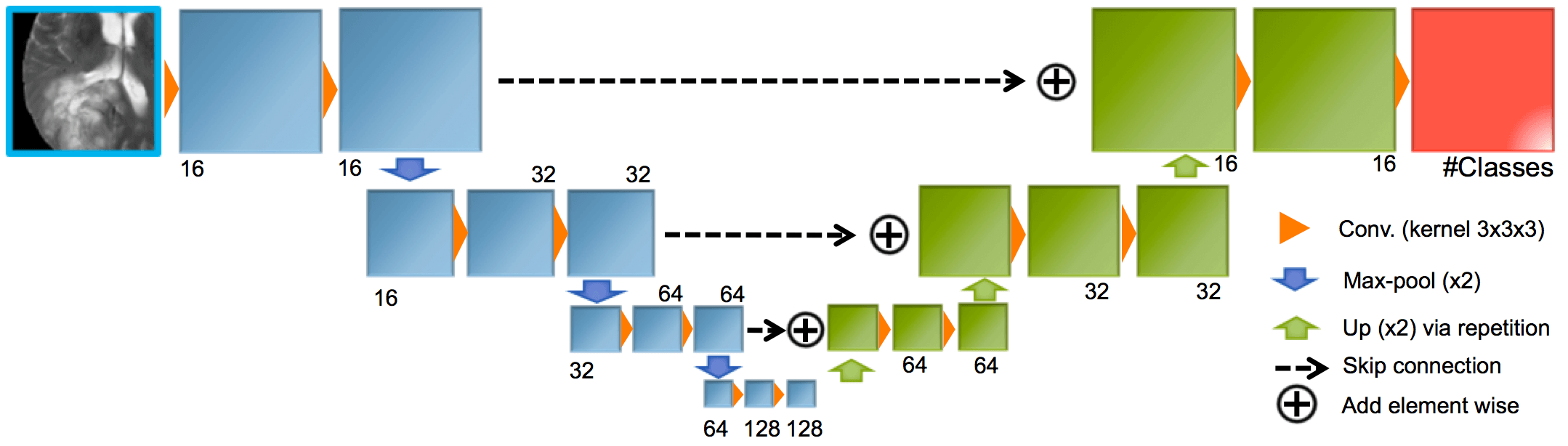}
\caption{Schematic of an adapted Unet used in our experiments. Depicted are number of feature maps per layer. All kernels and feature maps are 3D, even though not depicted for simplicity.}
\label{fig:unet}
\end{figure}

\noindent \textit{Model description:} We employ two 3D versions of the U-Net architecture \cite{ronneberger2015u} in our ensemble. The main elements of the first architecture are depicted in Fig.~\ref{fig:unet}. In this version we follow the strategy suggested in \cite{guerrero2017white} to reduce model complexity, where skip connections are implemented via summations of the signals in the up-sampling part of the network, instead of the concatenation originally used. The second architecture is similar but concatenates the skip connections and uses strided convolutions instead of max pooling. All layers use batch normalisation, ReLUs and zero-padding.

\noindent \textit{Training Details:}
The U-Nets were trained with input patches of size 64$\times$64$\times$64. The patches were sampled only from within the brain, with equal probability being centred around a voxel from each of the four labels. They were trained minimizing cross entropy via AdaDelta and Adam respectively, with different optimization, regularization and augmentation meta-parameters.

\subsection{Ensembling}

The above models are all trained completely separately. At testing time, each model segments individually an unseen image and outputs its class-confidence maps. The models are then ensembled into EMMA, according to eq.~\ref{eq:emmaFormula}. For this, the ensemble's confidence maps for each class are created by calculating for each voxel the average confidence of the individual models for the voxel to belong to this class. The final segmentation made by the EMMA is performed by assigning to each voxel the class with the highest confidence.

\subsection{Implementation details}

The original implementation of DeepMedic was used for the corresponding two models, along with the default meta-parameters, publicly available on \url{https://biomedia.doc.ic.ac.uk/software/deepmedic/}. The FCNs were implemented using DLTK, a deep learning library with a focus on medical imaging applications that allowed quick implementation and experimentation (\url{https://github.com/DLTK/DLTK}). Finally, an adaptation of the Unet will be released on \url{https://gitlab.com/eferrante}.


\section{Evaluation}
\label{sec:evaluation}

\subsection{Material}
Our system was evaluated on the data from the Brain Tumour Segmentation Challenge 2017 (BRATS) \cite{menze2015multimodal,bakasBrats17data1,bakasBrats17data2,bakasBrats17data3}. The training set consists of 210 cases with high grade glioma (HGG) and 75 cases with low grade glioma (LGG), for which manual segmentations are provided. The segmentations include the following tumour tissue labels: 1) necrotic core and non enhancing tumour, 2) oedema, 4) enhancing core. Label 3 is not used. The validation set consists of 46 cases, both HGG and LGG but the grade is not revealed. Reference segmentations for the validation set are hidden and evaluation is carried out via an online system that allows multiple submissions. In the testing phase of the competition, a test set of 146 cases is provided to the teams, and the teams have a 48 hours window for a single submission to the system. For evaluation, the 3 predicted labels are merged into different sets of whole tumour (all labels), the core (labels 1,4) and the enhancing tumour (label 4). For each subject, four MRI sequences are available, FLAIR, T1, T1 contrast enhanced (T1ce) and T2. The datasets are pre-processed by the organisers and provided as skull-stripped, registered to a common space and resampled to isotropic $1mm^3$ resolution. Dimensions of each volume are $240\times240\times155$.

\subsection{Preprocessing: Ensembling intensity normalisation methods}

\begin{figure}[t] 
\centering
\includegraphics[clip=true, trim=0pt 0pt 0pt 0pt, width=0.8\textwidth]{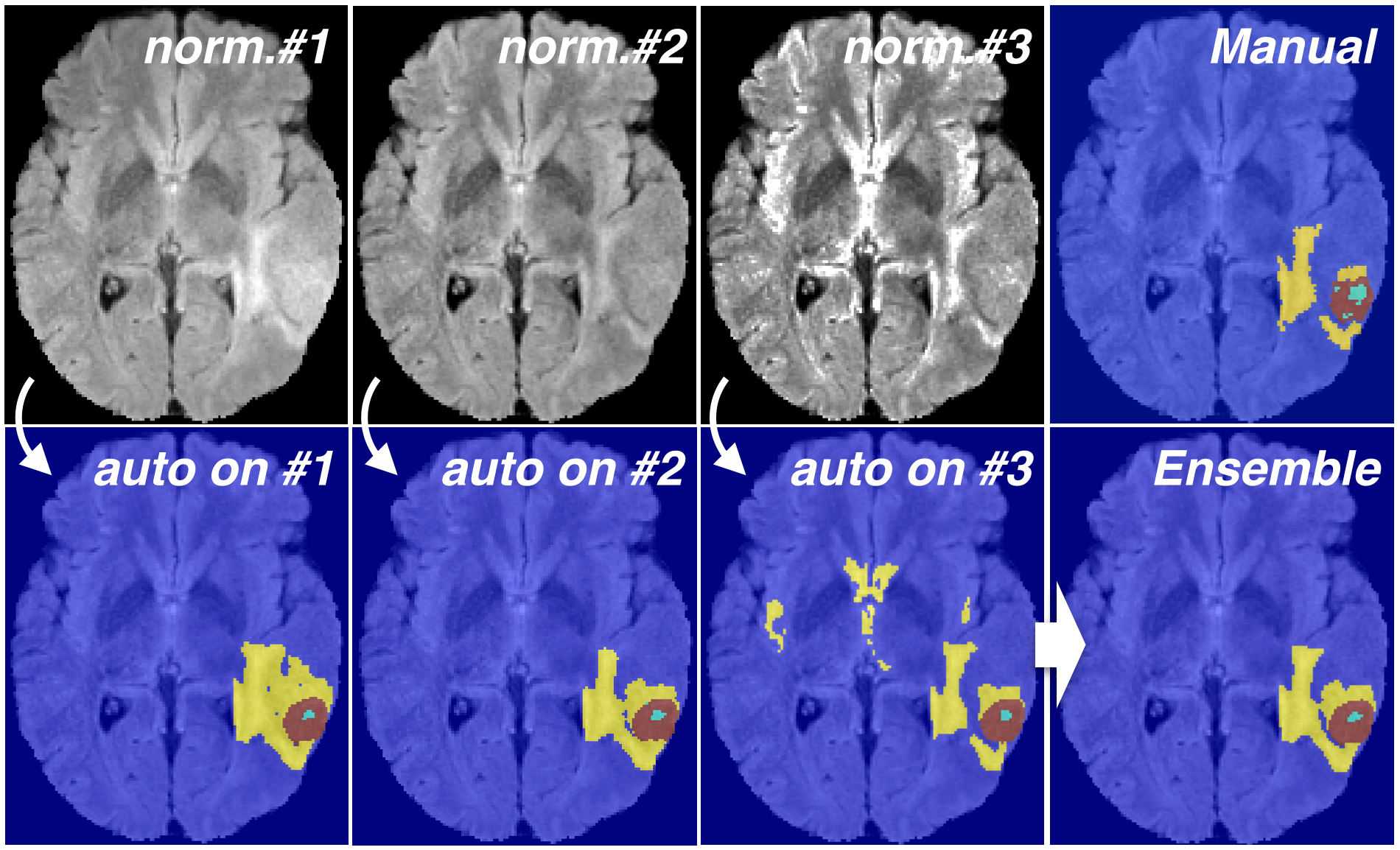}
\caption{Results are affected by normalization. To make a system robust to this factor, we introduce in EMMA models trained on differently normalized data.}
\label{fig:ensNorm}
\vspace{-5mm}
\end{figure}

We experimented with three different versions of intensity normalisation as pre-processing: 1) Z-score normalisation of each modality of each case individually, with the mean and stdev of the brain intensities. 2) Bias field correction followed by (1). 3) Bias field correction, followed by piece-wise linear normalisation \cite{nyul2000new}, followed by (1). Preliminary comparisons were inconclusive. We instead chose to average away the normalisation's effect with EMMA. Three instances of each network were trained, each on data processed with different normalisation. They were applied to correspondingly processed images for inference and all results were averaged in EMMA (Fig.~\ref{fig:ensNorm}).

\subsection{Results}
\label{subsec:results}

We provide the results that EMMA achieved on the validation and testing set of the BRATS'17 challenge\footnote{Leaderboard: \url{https://www.cbica.upenn.edu/BraTS17/lboardValidation.html}} on Table~\ref{table:results17}. Our system won the competition by achieving the overall best performance in the testing phase, based on Dice score (DSC) and Haussdorf distance. We also show results achieved on the validation set by the teams that ranked in the next two positions at the testing stage. No testing-phase metrics are available to us for these methods. We note that EMMA achieves similar levels of performance on validation and test sets, even though the latter contains data from different sources, indicating the robustness of the method. In comparison, competing methods were very good fits for the validation set, but did not manage to retain the same levels on the testing set. This emphasizes the importance of research towards robust and reliable systems.

\begin{table}[!h]
\centering
\scriptsize
\caption{Performance of EMMA on the validation and test sets of BRATS 2017 (submission id biomedia1). Our system achieved the top segmentation performance in the testing stage of the competition. For comparison we show the performance on validation set of the teams that ranked in the next two position. Performance of other teams in the testing stage is not available to us.}
\label{table:results17}
\begin{tabular}{@{}lcccccccccc@{}}
\toprule
              & \multicolumn{3}{c}{DSC}  & \multicolumn{3}{c}{Sensitivity} & \multicolumn{3}{c}{Hausdorff\_95}      \\ \cmidrule(lr){2-4} \cmidrule(lr){5-7} \cmidrule(lr){8-10} 
              	& Enh. & Whole 	& Core 		& Enh.   & Whole  	& Core  	& Enh. & Whole & Core & \#submits \\ \midrule

EMMA (val)	& 73.8	& 90.1	& 79.7	& 78.3	& 89.5	& 76.2	& 4.50 & 4.23 & 6.56 & 2 \\

UCL-TIG (val) & 78.6	& 90.5	& 83.8	& 77.1	& 91.5	& 82.2	& 3.28 & 3.89 & 6.48 & 21 \\

MIC\_DKFZ (val)	& 73.2	& 89.6	& 79.7	& 79.0	& 89.6	& 78.1	& 4.55 & 6.97 & 9.48 & 2\\ \hline

EMMA (test)	& 72.9	& 88.6	& 78.5	& -	& -	& -	& 36.0 & 5.01 & 23.1 & 1\\

\bottomrule
\end{tabular}
\end{table}


\section{Conclusion}
\label{sec:conclusion}

\begin{figure}[t] 
\centering
\includegraphics[clip=true, trim=0pt 0pt 0pt 0pt, width=1.\textwidth]{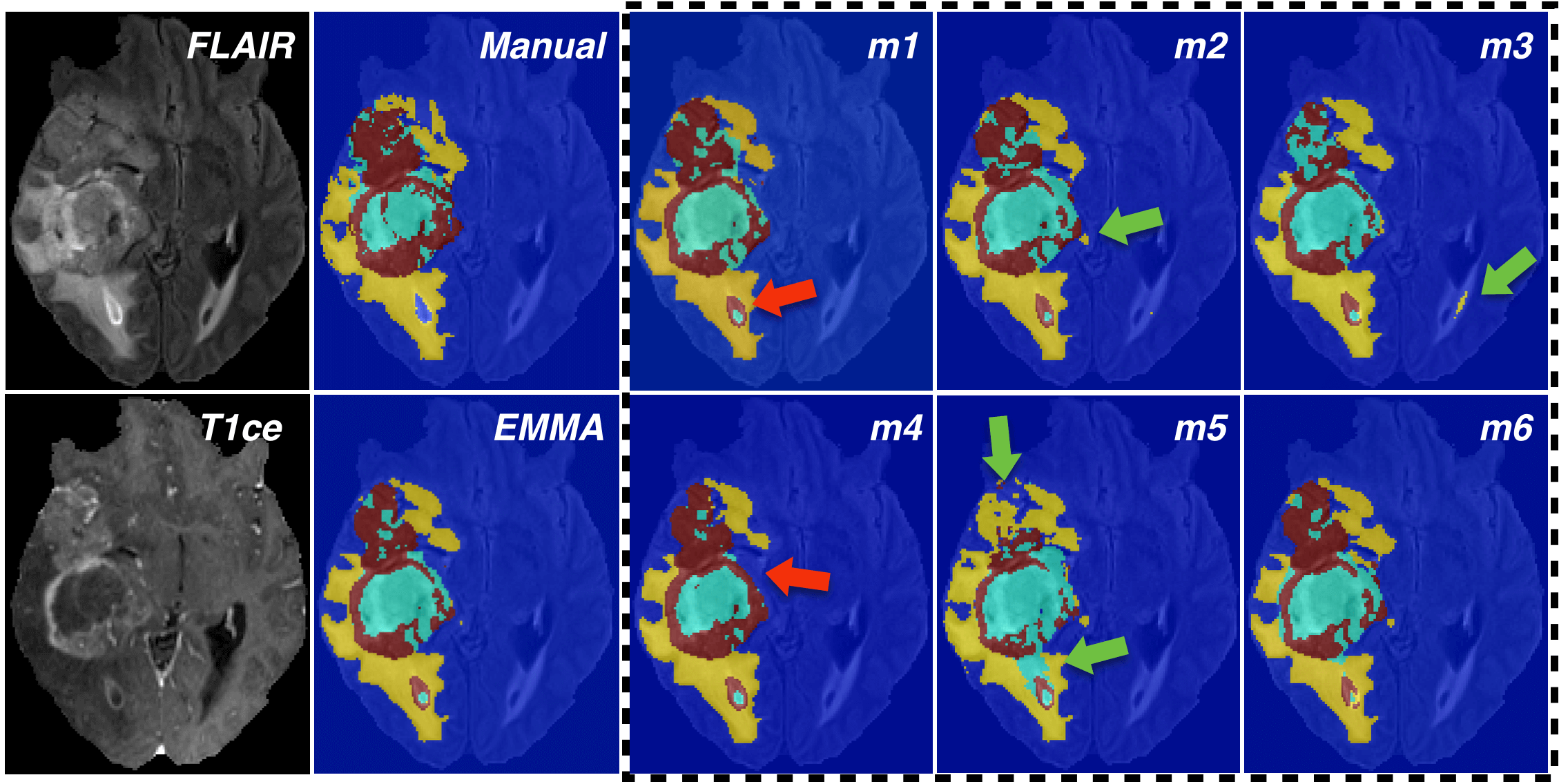}
\caption{FLAIR, T1ce and manual annotation of a case in the training set, along with automatic segmentation from preliminary version of EMMA consisting of six models. Green arrows point inconsistent mistakes by the individual model that are corrected by the ensembling, while red arrow shows a consistent mistake.}
\label{fig:emmaVisuals}
\end{figure}

Neural networks have been proven very potent, yet imperfect estimators, often making unpredictable errors. Biomedical applications are reliability-critical however. For this reason we first concentrate on improving robustness. Towards this goal we introduced EMMA, an ensemble of widely varying CNNs. By combining a heterogeneous collection of networks we construct a model that is insensitive to independent failures of CNN components and thus generalises well (Fig.~\ref{fig:emmaVisuals}). We also introduced the new perspective of ensembling for objectiveness. By marginalizing out via ensembling the biased behaviour introduced by configuration choices, EMMA is a model more fit for objective analysis. Even though the individual networks have straight-forward architectures and were not optimized for the task, EMMA won the first position in the final testing stage of BRATS 2017 competition among 50+ teams, indicating strong generalisation.

By being robust to suboptimal configurations of its components, EMMA may offer re-usability on different tasks, which we aim to explore in the future. EMMA could also be useful in unbiased investigation of factors such as sensitivity of CNNs to different sources of domain shift that is strongly affecting large-scale studies \cite{kamnitsas2017unsupervised}, or estimating amount of training data required for a task. Finally, EMMA's uncertainty could serve as a more objective measure of what type of patients or tumours are most challenging to learn.


\section{Acknowledgements}
\label{sec:acknowledgements}

This work is supported by the EPSRC (EP/N023668/1, EP/N024494/1 and EP/P001009/1) and partially funded under the 7th Framework Programme by the European Commission (CENTER-TBI: https://www.center-tbi.eu/). KK is supported by the President’s PhD Scholarship of Imperial College London. EF is beneficiary of an AXA Research Fund postdoctoral grant. NP is supported by Microsoft Research through its PhD Scholarship Programme and the EPSRC Centre for Doctoral Training in High Performance Embedded and Distributed Systems (HiPEDS, Grant Reference EP/L016796/1). We gratefully acknowledge the support of NVIDIA with the donation of GPUs for our research.

\bibliographystyle{splncs}
\bibliography{cites}

\end{document}